\documentclass{article}

\usepackage{arxiv}

\usepackage[utf8]{inputenc} 
\usepackage[T1]{fontenc}    
\usepackage{hyperref}       
\usepackage{url}            
\usepackage{booktabs}       
\usepackage{amsfonts}       
\usepackage{nicefrac}       
\usepackage{microtype}      
\usepackage{lipsum}		
\usepackage{graphicx}
\usepackage{natbib}
\usepackage{doi}

\title{ByteStorm: a multi-step data-driven approach for Tropical Cyclones detection and tracking}

\author{ 
    \href{https://orcid.org/0009-0001-0440-7962}{\includegraphics[scale=0.06]{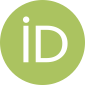}\hspace{1mm}Davide Donno\textsuperscript{1,2}} \\
    \And
    \href{https://orcid.org/0000-0002-9206-2385}{\includegraphics[scale=0.06]{orcid.png}\hspace{1mm}Donatello Elia\textsuperscript{1}} \\
    \And
    \href{https://orcid.org/0000-0002-2969-1517}{\includegraphics[scale=0.06]{orcid.png}\hspace{1mm}Gabriele Accarino\textsuperscript{1,3}} \\
    \And
    \href{https://orcid.org/0009-0003-6965-909X}{\includegraphics[scale=0.06]{orcid.png}\hspace{1mm}Marco De Carlo\textsuperscript{1}} \\
    \And
    \href{https://orcid.org/0000-0001-7987-4744}{\includegraphics[scale=0.06]{orcid.png}\hspace{1mm}Enrico Scoccimarro\textsuperscript{1}} \\
    \And
    \href{https://orcid.org/0000-0001-7777-8935}{\includegraphics[scale=0.06]{orcid.png}\hspace{1mm}Silvio Gualdi\textsuperscript{1}} \\
}

\date{}

\renewcommand{\headeright}{Preprint}

\hypersetup{
    pdftitle={ByteStorm: a multi-step data-driven approach for Tropical Cyclones detection and tracking},
    pdfsubject={},
    pdfauthor={Davide Donno, Donatello Elia, Gabriele Accarino, Marco De Carlo, Enrico Scoccimarro, Silvio Gualdi},
    pdfkeywords={Tropical Cyclones Tracking \and Deep Learning \and Computer Vision \and Atmospheric Science \and Machine Learning \and Multi-Object Tracking},
}

\begin{document}
\maketitle

1 CMCC Foundation - Euro-Mediterranean Center on Climate Change, Via Marco Biagi, 5, Lecce, Italy \\
2 University of Salento, Department of Engineering for Innovation, Via per Monteroni, Lecce, Italy \\
3 Department of Earth and Environmental Engineering, Columbia University, New York, NY, USA \\

\begin{abstract}
Accurate tropical cyclones (TCs) tracking represents a critical challenge in the context of weather and climate science. Traditional tracking schemes mainly rely on subjective thresholds, which may introduce biases in their skills on the geographical region of application and are often computationally and data-intensive, due to the management of a large number of variables. We present \textit{ByteStorm}, an efficient data-driven framework for reconstructing TC tracks. It leverages deep learning networks to detect TC centers (via classification and localization), using only relative vorticity (850 mb) and mean sea-level pressure. Then, detected centers are linked into TC tracks through the BYTE algorithm. \textit{ByteStorm} is benchmarked with state-of-the-art deterministic trackers on the main global TC formation basins. 
The proposed framework achieves good tracking skills in terms of Probability of Detection and False Alarm Rate, accurately reproduces Seasonal and Inter-Annual Variability, and reconstructs reliable, smooth and coherent TC tracks.
These results highlight the potential of integrating deep learning and computer vision to provide robust, computationally efficient and skillful data-driven alternatives to TC tracking.
\end{abstract}


\section{Introduction}
\label{sect:intro}

Tropical Cyclones (TCs) are among the most destructive natural phenomena, causing widespread damage and disruption globally \citep{kerry2003}. Their formation and development result from complex interactions between the ocean and atmosphere, modulated by large-scale circulation patterns. For the development of a TC, several conditions must occur \citep{gray1975,garner2023}: warm sea surface temperatures from which the TC draws energy due to the evaporation, the influence of the Coriolis force along with low wind shear and ample humidity and a pre-existing low-pressure disturbance.  
Every year, an average of 90 TCs occur worldwide \citep{emanuel_nolan2004}, and climate change is making them stronger and more destructive \citep{elsner2008,mendelsohn2012,sun2017}.
Owing to their significant socio-economic impacts \citep{wmo2023} and strong sensitivity to climate variability \citep{mueller2024}, the accurate monitoring and prediction of TCs remains a critical challenge in both weather and climate science. 

Detection and tracking of these phenomena in large climate datasets is an active area of research for the climate community \citep{scoccimarro2014,dabhade2021,garner2021,chand2022}. 

The identification of TCs in such datasets is traditionally performed using deterministic tracking algorithms, or TC trackers \citep{horn2014}, which detect TC structures within gridded climate fields, locate their centers, and link them across time, resulting in TC tracks \citep{bourdin_2022}.
Specifically, TC tracking schemes comprise two consecutive sub-tasks: \textit{detection} and \textit{linking}. The detection task aims to localize TC occurrences across space and time, and is typically highly permissive, as it detects a wide number of false positives and small disturbances. Then, the linking task joins previously detected TC occurrences across time based on some physical constraints (e.g., TC eye located in the local minima of mean sea level pressure, maximum TC distance after 6 hours etc.), further removing most of these unwanted detections.

Traditional tracking schemes are generally classified into \textit{physics-based} \citep{camargo2002,chauvin_2006,zhao2009,horn2014,murakami2014,zarzycki2017} and \textit{dynamics-based} \citep{hodges_2017,strachan2013,tory2013b}. Physics-based trackers rely on thermodynamic variables, typically identifying local minima in the sea level pressure and confirming the presence of a warm-core using surface temperature anomalies or geopotential thickness. They often apply an additional intensity-based criterion applied on surface wind speed or vorticity to validate detections. In contrast, dynamics-based trackers focus on vorticity fields and the velocity derivatives to detect the TC centers \citep{bourdin_2022}. 

The aforementioned methods typically rely on threshold-based criteria, which makes them sensitive to expert's parameters selection and introduces potential subjectivity in their design \citep{dabhade2021,enz2022}. Moreover, these thresholds often vary with geographical region and TC category \citep{befort2020,bloemendaal2021}, leading to systematic biases and limiting the generalizability of the algorithms to regions or datasets beyond those used for calibration. 
In addition, TC trackers, such as those described in \cite{hodges_2017} and \cite{horn2014}, are often computationally and data-intensive, requiring the storage and management of a considerable number of variables, at high temporal resolutions, along with extensive post-processing and execution time. As a result, the analysis of the climatic characteristics of TCs remains a resource-demanding and time-consuming task, hindering large-scale and long-term studies.

In this landscape, Machine Learning (ML) opened new avenues across key areas of climate science, including the analysis of TCs and their effect on the environment.
A consistent part of these works frames TC detection and linking as an Object Detection (OD) task. Mask- and Faster- Region-based Convolutional Neural Networks have been employed in \cite{wang2020,xie2020,nair2022}, whereas the You Only Look Once architecture has been used in \cite{shakya2020,pang2021,haque2022,lam2023}. 
\cite{kumlerbonfanti2020} explored U-Net-based segmentation approaches for accurate TC detection, whereas \cite{kim2019a} compared the skills of Decision Trees, Random Forest, Support Vector Machines, and Linear Discriminant Analysis for the TC classification. All the aforementioned works leverage satellite observations for training.
Other studies explored TC detection with different data sources: \cite{matsuoka2018} trained a Convolutional Neural Network (CNN) for binary classification of TCs and their precursors using Outgoing Long-wave Radiation from Cloud-resolving Global Atmospheric Simulation, whereas \cite{galea2023} proposed TCDetect, a data-driven framework for detecting TC presence and absence leveraging ERA-Interim \cite{dee2011} dataset for the climate variables and an online data-reduction method.
Although their advances, most of these approaches consist in TC detection or binary classification, limiting their use to climate applications.

Another line of research focuses its attention to TC tracking through data-driven approaches. 
For example, \cite{yan2023} defines the TROPHY framework to quantify the structural stability of TCs exploiting wind fields, making it suitable for TC tracking. Despite the innovative physics-informed approach, the algorithm doesn't computationally scale well with the number of TCs. 
In a more recent work, \cite{lin_2025} proposed a Bayesian Inference and Dynamic Programming Tracking (BIDTrack) method. The algorithm is optimized through a Bayesian Interval Optimization process, refining parameters to keep only statistically significant and physically consistent TC candidates, resulting in faithful TC tracks.
\cite{ayar_2025} exploited an Ensemble Random Forest (ERF) model composed of 100 elements. The ERF is designed as a classification task, detecting TC centers and then aligning them in time with a simple tracking algorithm, achieving very good TC tracking skills.

In \cite{accarino_2023} we proposed an hybrid approach, where 13 Deep Learning (DL) models were used in an ensemble setting to provide estimates of TC center locations, which were subsequently linked together into plausible tracks by means of a deterministic tracking scheme, following \cite{zhao2009} and \cite{scoccimarro2017}. Although our approach showed promising results compared to deterministic trackers in terms of detection accuracy and tracks reconstruction, several limitations emerged. First, the use of 6 environmental predictors combined with an ensemble of 13 models resulted in a computationally demanding pipeline, limiting the scalability of the method when applied to large datasets. Second, the tracking component remained deterministic and still based on thresholds to connect candidate TC centers across time steps, introducing a degree of subjectivity in the resulting tracks. Finally, the methodology was developed and tested specifically on the North Pacific and Atlantic basins, which restricted its applicability to a regional context. These aspects highlighted the need for a more efficient, fully data-driven, and global approach.

Building on the promising results of our previous work, we present \textit{ByteStorm}, a novel framework that combines two DL models - for TC classification and localization - with BYTE \citep{zhang_2022}, a state-of-the-art computer vision Multi-Object Tracking (MOT) algorithm for TC track reconstruction. \textit{ByteStorm} links individual TC centers into temporally coherent tracks, thus providing and end-to-end, scalable solution for storm identification and evolution tracking.

Unlike traditional tracking approaches ByteStorm adopts a fully data-driven paradigm that eliminates the need for subjective parameter choices during the tracking phase. The framework bridges advances in DL-based pattern recognition with modern computer vision object tracking algorithms, enabling a more flexible and robust treatment of TC evolution in gridded environmental data. At the same time ByteStorm simplifies the input requirements of the detection pipeline: it relies on only two environmental predictors, drastically reducing data volume and computational cost, while maintaining strong detection and tracking skills. We evaluate the capabilities of ByteStorm by benchmarking it against four state-of-the-art deterministic tracking schemes on global TC basins. We adopt multiple quantitative metrics and we examine ByteStorm's skills on different datasets and selected historical TC case studies. 

The remainder of the paper is organized as follows. Section \ref{sect:materials} describes the dataset generation and labeling process. Section \ref{sect:methods} outlines the experimental setup and the evaluation metrics used to assess model performance. 
Section \ref{sect:results} presents an assessment of ByteStorm skills, while Section \ref{sect:discussion} discusses and contextualizes the findings. Finally, Section \ref{sect:conclusions} draws the conclusions and outlines potential directions for future research.

\section{Materials}
\label{sect:materials}

In this study, from a geographical perspective, we focus on the most relevant global TC formation basins, namely North Pacific comprising Eastern and Western basins (ENP, WNP), South Pacific (SP), North and South Indian (NI, SI) and North Atlantic (NATL), defined as in \cite{knutson2020}\footnote{Supplementary material to the paper}. Due to the very low amount of historical TC records, we exclude South Atlantic basin from the analysis.

\subsection{Data Sources}
\label{sect:data_sources}

\subsubsection{International Best Tracks Archive for Climate Stewardship}
The International Best Tracks Archive for Climate Stewardship (IBTrACS) \citep{knapp2010} provides the most accurate global archive of historical TC records. It integrates observations from 12 meteorological agencies at a 3-hourly temporal resolution and is gridded at a spatial resolution of $0.1^{\circ} \times 0.1^{\circ}$. The dataset also includes TC intensity metrics (e.g., wind speed), observation timestamps, and metadata such as the nature of the storm (e.g., Tropical, Extra Tropical, Disturbance Storm, etc.). Although IBTrACS extends back to 1841, in this study we focus on the 1980-2022 period, discarding earlier records due to limited satellite coverage and excluding most recent entries that may still undergo post-processing and reanalyses, potentially compromising data reliability \citep{ibtracs_doc2019}.

\subsubsection{ERA5 Reanalysis}
The ERA5 reanalysis dataset \citep{hersbach2023a} includes key climate variables that are essential for understanding TC formation. Climate variables are provided on a global regular grid at a spatial resolution of $0.25^{\circ} \times 0.25^{\circ}$, corresponding to $721\times1440$ (H $\times$ W) grid points. To ensure consistency, we limit ERA5 data to the same period as IBTrACS (1980-2022). After a preliminary evaluation aimed at reducing the number of variables considered in our approach, from ERA5 dataset we select \textit{relative vorticity at 850mb} (RV850) and \textit{mean sea level pressure} (MSLP) among the key variables typically associated with TC events.

\subsubsection{JRA-3Q Reanalysis}
To assess the extrapolation skills on a different dataset, we also considered the high-resolution Japanese Reanalysis for Three Quarters of a Century (JRA-3Q) \citep{jra3q2023}. JRA-3Q is the third reanalysis developed from the Japanese Meteorological Agency (JMA). 
JRA-3Q provides improved representation of TCs using a method that generates fictional TCs to tackle the artificial weakening of TCs present in the previous reanalysis \citep{kosaka2024}.
This dataset comes at a 6-hourly temporal resolution and a spatial resolution of $ 0.375^\circ \times 0.375^\circ$, that corresponds to a $480 \times 960$ ($H \times W$) grid points. To ensure compatibility with ERA5 grid, we remap the JRA-3Q data from the original grid structure and resolution to the ERA5 one using a conservative method. From JRA-3Q we extract the same variables used for ERA5, namely RV850 and MSLP, covering a recent three-year period (2020–2022).  

\subsection{Dataset Processing}
\label{sect:data_prep_aug}

\subsubsection{Data Filtering}
\label{sect:ibtracs_filtering}

Since 6-hourly samples of IBTrACS dataset are provided with more accurate information regarding TC characteristics than 3-hourly ones \citep{knapp2010}, we sample IBTrACS, ERA5 and JRA-3Q datasets at 6-hour intervals, at 00.00, 06.00, 12.00 and 18.00 UTC.
Furthermore, we spatially crop the dataset over the global basins, covering the region $0-360^{\circ}E$ and $70^{\circ}S-70^{\circ}N$.

We retain from IBTrACS only tracks labeled as \textit{main} in the "track type" field, while we discard tracks marked as \textit{provisional}, \textit{spur}, or \textit{provisional-spur} due to their lower reliability \citep{ibtracs_doc2019}. We consider all available storm natures in the analysis: \textit{Not Reported} (NR), \textit{Mixture} (MX), \textit{Disturbance Storm} (DS), \textit{Tropical Storm} (TS), \textit{Extra Tropical} (ET) and \textit{Sub Tropical} (SS). Based on the filtered IBTrACS data, we extract ERA5 maps including only time steps when a TC is recorded within the selected TC formation basins.

\subsubsection{Patches Generation and Datasets Creation}
\label{sect:patch_gen_dataset}

The filtering process results in a total of 55 215 maps. We crop each ERA5 map, with dimensions of $560 \times 1440$ grid points and containing the two selected variables (RV850 and MSLP), in correspondence of each basin, resulting in 6 different cropped sub-maps. 
Furthermore, we divide each sub-map into non-overlapping patches of size $40 \times 40$. 
To match spatial resolutions, we interpolate TC center coordinates from IBTrACS - originally at $0.1^{\circ} \times 0.1^{\circ}$ resolution - to match the $0.25^{\circ} \times 0.25^{\circ}$ ERA5 grid. We then assign each patch a label based on the presence of a TC center: label $1$ for patches containing a TC center (i.e., \textit{positive patches}) and label $0$ for patches that do not (i.e., \textit{negative patches}). 
For each \textit{positive patch}, since the TC center can appear anywhere within the patch, we also record the \textit{(row, column)} coordinates of the TC center within the $40 \times 40$ grid (each in the range $[0, 39]$).

To train the ByteStorm DL models, we split the dataset as follows. Using a periodic sampling approach, we construct the test set by selecting a month every five ($k=5$ step size) from January 1980 to December 2019, covering a 40-year period. This systematic sampling ensures a uniform representation of all months, accounting for the seasonal nature of TCs, and provides an independent and representative test set. By spanning four decades, the test set also allows the model to be robust to evolving climate conditions associated with climate change, that may affect the environmental variables used.
The remaining months are divided into a training set (1980-2009) and a validation set (2010-2019). The test set is used to benchmark ByteStorm against deterministic TC trackers (see Section \ref{sect:res_comp_trackers}) and to provide statistically meaningful out-of-sample performance metrics. To assess extrapolation capabilities, we apply the same patch-extraction procedure on recent years to ERA5 and JRA-3Q data (2020–2022), generating additional independent test samples.

To construct the training and validation datasets, for each basin we adopt a patch-selection strategy based on the following criteria:

\begin{itemize}

    \item \textit{\textbf{Cyclone Patches}} (total: 181 016): patches that \textbf{contain} a TC center. It is important to note that the TC eye can occur at any \textit{(row, column)} location of the patch, meaning the storm may be only partially visible depending on its position.
    
    \item \textit{\textbf{Nearest Patches}} (total: 499 953): patches that \textbf{do not contain} a TC center but selected as the three corner-adjacent patches closest to each corresponding \textit{cyclone patch}. This selection enriches the dataset with negative samples that may still exhibit TC-related patterns, thus improving the model's ability to generalize. By design the number of \textit{nearest patches} is approximately three times that of \textit{cyclone patches}.
    
    \item \textit{\textbf{Random Patches}} (total: 181 016): additional \textbf{non-cyclone} patches randomly sampled from the remaining patches. 
    One random patch is selected per \textit{cyclone patch}, ensuring numerical balance.
    
\end{itemize}

This sampling strategy results in a \textit{positive-to-negative} patch ratio of approximately 1:4.

\section{Methods}
\label{sect:methods}

\subsection{Experimental Setup}
\label{sect:experimental_setup}

The experimental setup of ByteStorm consists of two main steps: 
\begin{enumerate}
    \item Two DL models, (i.e., \textit{classification}, \textit{localization}) are jointly used to detect TC center locations;
    \item The BYTE MOT algorithm \citep{zhang_2022} is used to align TC center detections over time, thus reconstructing full TC tracks.
\end{enumerate}

\begin{figure}
    \centering
    \includegraphics[width=\textwidth]{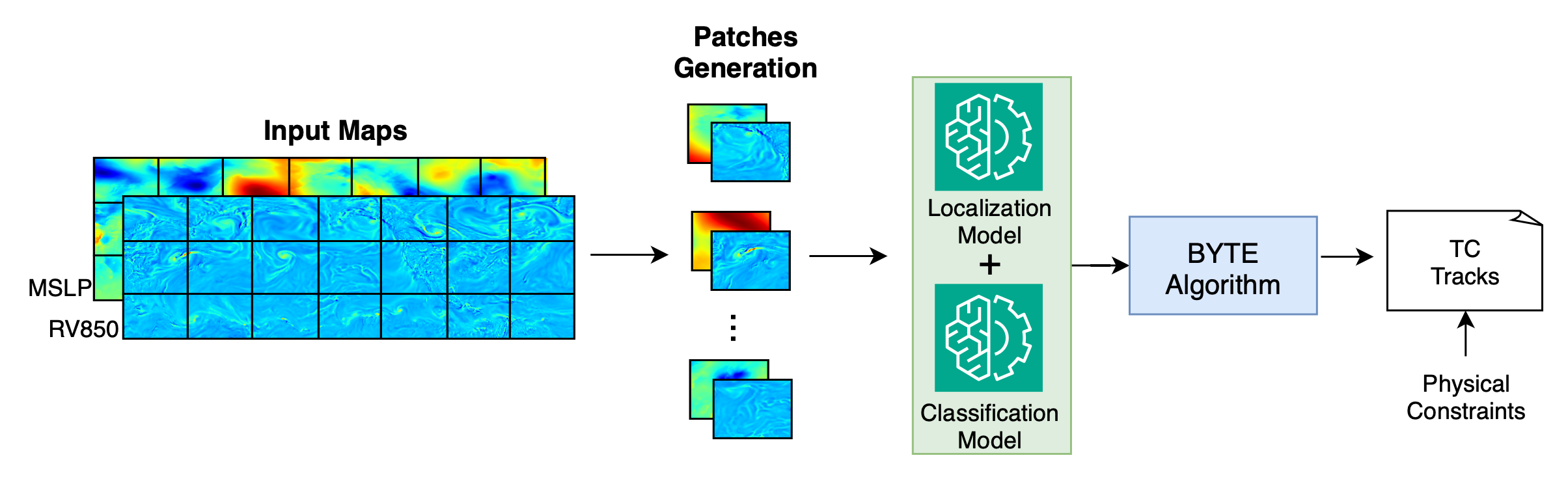}
    \caption{
    Overview of the ByteStorm approach. The input maps, consisting of RV850 and MSLP predictors, are divided into non-overlapping patches and processed by two models to (i) detect the TCs and (ii) locate their position within the patch. The BYTE method is then applied to link TC center detections into coherent spatio-temporal tracks.
    }
    \label{fig:algo}
\end{figure}

In particular, the \textit{classification} model assigns a probability of TC occurrence to each patch, and, if a patch is classified as containing a TC phenomena (probability $\geq 0.5$, where $0.5$ represents the canonical logistic threshold), the \textit{localization} model estimates the TC center coordinates within that patch. 
Figure \ref{fig:algo} shows a high-level overview of the ByteStorm TC tracking algorithm. It first partitions the input predictors into non-overlapping patches (see Section \ref{sect:patch_gen_dataset}) and feeds them into the \textit{classification} model. If the classifier determines the presence of a TC within the patch, the \textit{localization} model is used to estimate its TC center location. By applying the two DL models on a set of consecutive 6-hourly maps, a wide number of TC center locations is detected. Finally, ByteStorm adopts the BYTE algorithm to link the TC centers into spatio-temporally plausible tracks.

\subsubsection{TC Detection: Classification and Localization DL models}
\label{sect:detection}

In this work, we decompose the TC detection task into two DL sub-tasks, namely \textit{classification} and \textit{localization}, for which two separate DL models are trained. To account for the physical differences among north- and south- hemispheres, we tailor two models (i.e., classification and localization) for each hemisphere. Therefore, for NI, WNP, ENP and NATL basins we use the northern DL models, whereas for SI and SP basins we use the southern DL models. At test time, we transparently apply the correct couple of models according to the selected hemisphere.
Classification and localization models both leverage the same Visual Geometry Group (VGG) architecture \citep{simonyan_2015}. The two physical predictors within each patch (i.e., RV850 and MSLP) are treated as 2-dimensional images stacked along the channel dimension, resulting in an input tensor of size $2 \times 40 \times 40$ ($C \times H \times W$). The VGG networks process these tensors as follows:

\begin{itemize}

    \item \textbf{Classification Model} - learns the relationship between input predictors and the presence of a potential TC and estimates the probability of TC occurrence through a sigmoid activation function in the output layer. We train the model using the dataset described in Section \ref{sect:patch_gen_dataset}.
    We use the Binary Cross Entropy as loss function.

    \item \textbf{Localization Model} - learns to identify TC-related spatial patterns and estimate the coordinates of the TC center within the $40 \times 40$ patch, expressed as \textit{(row, col)}. We train the model exclusively on the \textit{cyclone patches} using the Mean Absolute Error as loss function.

\end{itemize}

\begin{figure}
    \centering
    \includegraphics[width=\textwidth]{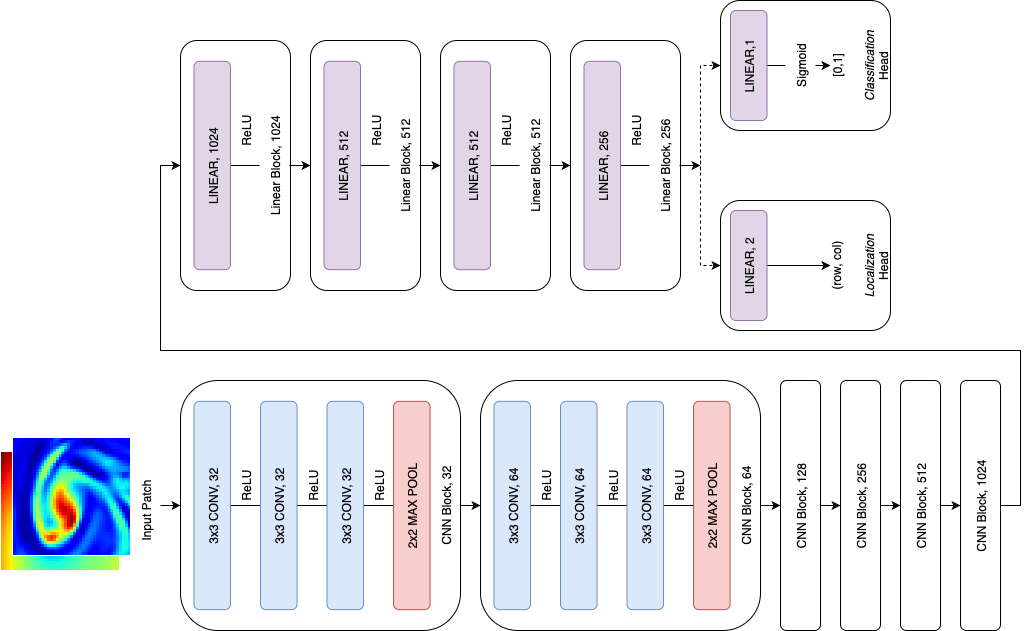}
    \caption{Architecture of the VGG-like models used in this work. A series of 6 CNN blocks is used as convolutional backbone. The CNN model is followed by a series of 4 Linear blocks that decode the spatial information coming from the CNN into a representation of TC presence/absence in the case of \textit{Classification} model or the location of the TC eye in the case of \textit{Localization} model. $f(x)$ is the ReLU activation function.}
    \label{fig:arch}
\end{figure}

As shown in Figure \ref{fig:arch}, the two VGG networks have a backbone composed of 6 CNN \citep{LeCun_1998} Blocks followed by 4 Linear Blocks and the output head. Each of the CNN Blocks comprises 3 convolutional layers with $3 \times 3$ kernel size, interleaved with the ReLU activation function and ending with a Max Pooling layer to halve the spatial image size. The number of filters doubles at each block, from 32 to 1024. The CNN Blocks ensure to capture complex spatial non-linear patterns related to TC occurrence, processing the multi-scale interaction between the two physical predictors directly in the latent space. The Linear Blocks, instead are composed of Linear layers interleaved with ReLU activation function. These blocks learn to translate the spatial information coming from the CNN into an encoded vector containing the information related to the TC position in the case of \textit{localization} model, or the TC presence in the case of \textit{classification} model. The number of filters in this case is 1 024, 512, 512, 256. After the CNN and Linear blocks, we apply a model-specific output layer. \textit{Classification} model has a scalar value in output after a sigmoid activation function that constrains the values between $0$ and $1$. \textit{Localization} model has two values in output, to predict a (\textit{row,col}) coordinate. Additionally, whereas the \textit{Classification} model is initialized with default weights, the \textit{Localization} one is initialized with Normal distribution with $0$ mean and $0.03$ standard deviation. This operation increased the magnitude of the weights, enabling the \textit{Localization} model to output values in the range $[0,39]$. We perform training using the AdamW optimizer \citep{loshchilov_2019}, with a batch size of 8 192 for 150 epochs.

Based on our experiments, considering the relatively small dataset, this $11M$ parameters' network configuration is balanced enough to ensure a good trade-off between bias and variance, thus avoiding both over- and under-fitting problems.

\subsubsection{Tracking Algorithm: BYTE}
\label{sect:tracking_algo}

BYTE is a simple and effective data-association algorithm widely used in the Computer Vision domain for MOT \citep{zhang_2022}. In contrast to other data-association algorithms, BYTE retains almost every detection box, separating them in low- and high-score ones. 

The algorithm can be tuned through several hyper-parameters: the \textit{Track Buffer} defines the maximum number of frames (i.e., 6-hourly time-steps in our case) that can miss after which a \textit{lost} track is labeled as \textit{removed}; \textit{Track Threshold} sets the confidence upper bound to consider a detection as valid; and the \textit{Match Threshold} defines the minimum Intersection-over-Union (IoU) overlap required to join two consecutive detections to the same track.

In this work, TC centers predicted by the \textit{localization} model are labeled with the probability score estimated by the \textit{classification} model and enclosed in a bounding box of size $31 \times 31$ pixels. The boxed detections are then processed by BYTE to reconstruct the TC tracks. To ensure physical consistency, we filter BYTE's outputs with additional physical constraints. We discard detections originated above $30^{\circ}$ of latitude, similarly to other state-of-the-art TC trackers \citep{hodges_2017}. Furthermore, following \cite{scoccimarro2017} and \cite{zhao2009}, we set the bounding box size and other hyper-parameters to enforce maximum TC displacements of $400$ km between consecutive 6-hourly time-steps, and discard all the tracks lasting less than 3 days (i.e., equivalent to 12 6-hourly time-steps). Appendix \ref{sect:appendix_a} provides full details on BYTE's parameters optimization.

\subsection{Evaluation Metrics}
\label{sect:val_metrics}

To benchmark ByteStorm's skills with the other trackers, we adopt several evaluation metrics: \textit{Probability Of Detection} (POD), \textit{False Alarm Rate} (FAR), \textit{Track Duration}, \textit{Seasonal} and \textit{Inter-Annual Variability} (IAV) and \textit{Track Smoothness}. 
To compute the aforementioned metrics we apply a matching algorithm between IBTrACS observations and the TC trackers' detections. 
We refer to \cite{bourdin_2022} for a deeper explanation of the matching procedure. 

\textbf{POD and FAR} are commonly used metrics to evaluate the performance of TC trackers. The POD measures the fraction of observed TCs correctly identified by the tracker, with higher values indicating better detection skill. The FAR quantifies the fraction of predicted TCs that do not correspond to observed storms, with lower values indicating fewer spurious detections. These metrics are computed as follows:

\begin{equation}
\label{eqn:pod}
POD = \frac{H}{H + M}
\end{equation}

\begin{equation}
\label{eqn:far}
FAR = \frac{FA}{H + FA}
\end{equation}

Where H (Hit) is the number of observed TC tracks that are correctly matched to at least one detected TC track during the TC lifetime, M (Miss) is the number of observed track with no corresponding detected TC track, while FA (False Alarms) is the number of detected TC tracks that do not correspond to any observed one.

\textbf{Seasonal Variability} describes the variation of the number of TCs across each month, giving insights about the seasonal cycle. We compute the Seasonal Variability by counting the total number of TC tracks detected/observed within the test set.

\textbf{IAV}: describes the year-to-year fluctuations in TC frequency. IAV represents the number of TC tracks detected within the test set months of each year.
To determine the quality of the reconstructed IAV, we first de-trend the TC tracker's and IBTrACS' IAVs and compute the Pearson correlation coefficient between the two curves.

\textbf{Track Duration}: measures the distribution of TC track lengths (in days) and serves to assess whether trackers correctly capture the temporal extent of cyclones. It highlights potential biases in underestimating or overestimating the duration of TC events. 

\textbf{Track Smoothness}: measures the variation of direction of a TC track, giving information about the shape of the track. We compute it as the standard deviation of the bearing angle variation (see below) between each consecutive detection. The lower the standard deviation is, the lower the TC centers change direction over time. Instead, if the smoothness value is high, it likely means that the detected TC track is noisier. Ideally, the predicted TC track should be as smooth as the variation shown in the IBTrACS ground truth.

Given a set of latitude and longitude coordinates $(\phi_i, \lambda_i)$ for $i = 1, \dots, N$, the bearing between points can be computed as in Bearing Angle Computation \citep{veness2022}.

\begin{equation}
\label{eqn:theta}
\theta_i = \tan^{-1} (\sin( \Delta \lambda_i) \cos(\phi_{i+1}), \cos(\phi_i) \sin(\phi_{i+1}) - \sin(\phi_i) \cos(\phi_{i+1}) \cos(\Delta\lambda_i) )
\end{equation}

where $\Delta\lambda_i = \lambda_{i+1} - \lambda_i$. Then the variation between successive bearings is: 

\begin{equation}
\label{eqn:dtheta}
\Delta\theta_i = \min ( |\theta_{i+1} - \theta_i|,\ 2\pi - |\theta_{i+1} - \theta_i| )
\end{equation}

Hence, the Track Smoothness score is defined as the standard deviation $\sigma$ of the angular variations:

\begin{equation}
\label{eqn:smoothness}
\sigma_{smoothness} = \sqrt{ \frac{1}{N-2} \sum_{i=1}^{N-2} (\Delta\theta_i - \bar{\Delta\theta})^2 }
\end{equation}

\section{Results}
\label{sect:results}

\subsection{Benchmark against Deterministic TC trackers}
\label{sect:res_comp_trackers}

We benchmark ByteStorm with four well-established deterministic trackers through a basin-wise evaluation, highlighting the key characteristics of the DL-based approach. Following the framework of \cite{bourdin_2022}, we select two physics-based trackers: CNRM \citep{chauvin_2006} and UZ \citep{ullrich_2021}, and two dynamic-based trackers: OWZ \citep{tory_2013} and TRACK \citep{hodges_2017}. These methods differ in terms of the variables considered, threshold criteria, and methodologies used for TC detection and track reconstruction. As a result, they can produce substantially different outcomes in terms of detected TC and FAs, offering a robust reference framework for evaluating ByteStorm.

We evaluate ByteStorm tracking skills against the filtered IBTrACS dataset described in Section \ref{sect:ibtracs_filtering}. To provide a more comprehensive assessment, the ground truth also includes less well-formed TC tracks, specifically TC records labeled as NR, MX, and DS (see Section \ref{sect:ibtracs_filtering}) in the IBTrACS documentation \citep{ibtracs_doc2019}. 

\subsubsection{Probability Of Detection and False Alarm Rate}
\label{sect:pod_far}

\begin{figure}
    \centering
    \includegraphics[width=\textwidth]{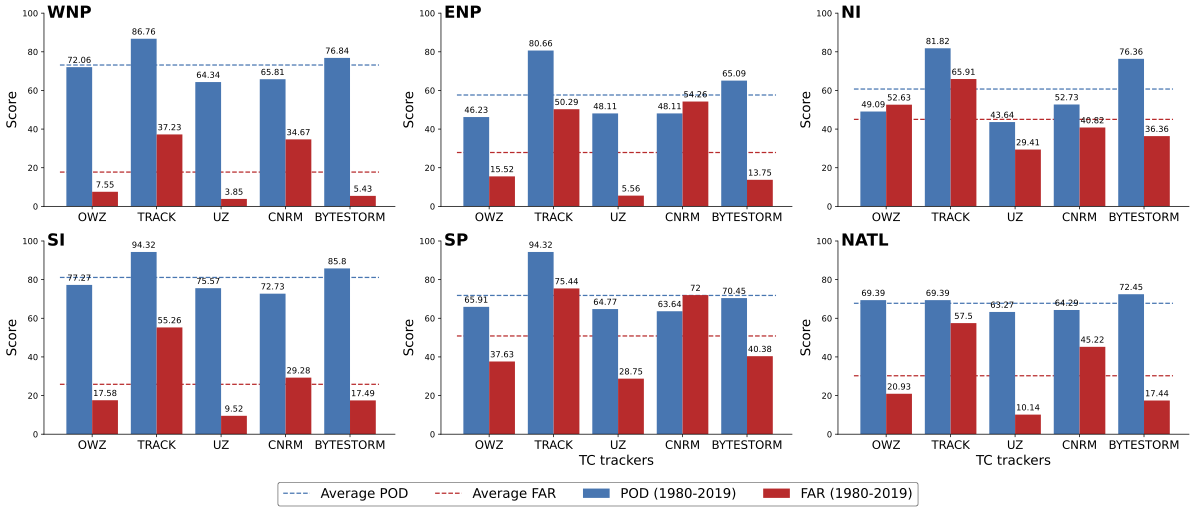}
    \caption{Probability of Detection (POD) - higher is better - and False Alarm Rate (FAR) - lower is better - of the deterministic TC trackers compared with ByteStorm, across the different TC formation basins.}
    \label{fig:pod_far}
\end{figure}

Figure \ref{fig:pod_far} compares the performance of ByteStorm with four deterministic trackers across the considered basins. Overall, ByteStorm demonstrates a favorable POD-FAR trade-off globally, achieving the second-highest POD after TRACK while maintaining one of the lowest FAR values, comparable to OWZ and UZ. 
Its strongest performance is observed in the WNP and SI basins, with POD of $76.84 \%$ and $85.8 \%$, and FAR of $5.43 \%$ and $17.49 \%$, respectively. As with most deterministic trackers, ByteStorm exhibits higher FAR in the NI and SP basins ($36.36\%$ and $40.38\%$), reflecting similar spurious detections. Despite heterogeneous skills across basins, these results demonstrate that ByteStorm is capable of capturing the majority of observed TC tracks within the test set. 

\begin{figure}
    \centering
    \includegraphics[width=\textwidth]{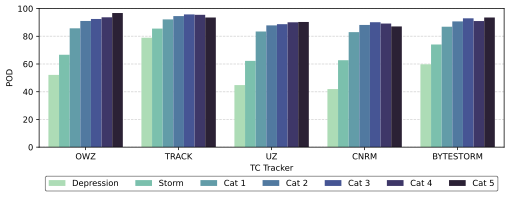}
    \caption{
    Probability of Detection (POD) divided by storm category according to the Saffir-Simpson scale. On the x-axis the TC tracker is reported. Each bar color represents the storm category.
    }
    \label{fig:pod_by_cat}
\end{figure}

Furthermore, we partition POD according to storm category, using the Saffir-Simpson scale (SSHS).
Figure \ref{fig:pod_by_cat}, shows ByteStorm's and the deterministic trackers' global POD by storm category. Similarly to the other trackers, ByteStorm's detection performance is lower for Tropical Depressions ($59.78\%$) and weak Tropical Storms ($74.07\%$), increasing with storm intensity up to $93.55\%$ for Category 5 storms. 
This increase reflects a physical expectation: more organized TCs, with sharper MSLP gradients and well-structured radial RV850 fields, are detected with higher confidence. Conversely, small, asymmetric or not well-organized systems remains more challenging. 

\begin{table}[]
\centering
\begin{tabular}{@{}lllllll}
\toprule
& Basin                 & OWZ & TRACK & UZ & CNRM & BYTESTORM\\ 
\midrule
& WNP  &     \textbf{0.82} &0.54 &0.71 &0.56 & 0.79 \\
\hline
& ENP  &     0.54 &0.58 &0.66 &0.39 &\textbf{0.73} \\
\hline
& NI   &     0.4  & 0.61 &0.57 &0.57 &\textbf{0.66} \\
\hline
& SI   &     0.79 &0.84 &0.85 &0.85 &\textbf{0.91} \\
\hline
& SP   &    0.78 &0.64 &0.8  &0.71 &\textbf{0.81} \\
\hline
& NATL &     \textbf{0.86} &0.52 &\textbf{0.86} &0.68 &0.8 \\
\end{tabular}
\caption{
    Pearson correlation coefficient computed on the de-trended Inter-Annual Variability for each deterministic tracker and ByteStorm. Metrics in bold indicate the best-performing model.
}
\label{tab:pearson_iav}
\end{table}

\subsubsection{Seasonal and Inter-annual Variability}
\label{sect:interannual_Var}

\begin{figure}
    \centering
    \includegraphics[width=\textwidth]{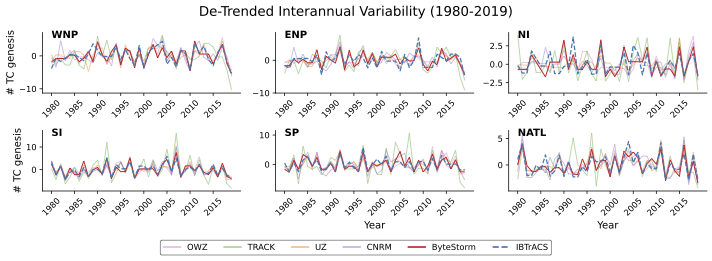}
    \caption{De-trended inter-annual variability of deterministic TC trackers compared to ByteStorm across the different TC formation basins. ByteStorm is reported as a red thicker line, while IBTrACS ground truth is represented as a dashed blue line.}
    \label{fig:iav}
\end{figure}

In Figure \ref{fig:iav}, we show the IAV of TC activity across the different trackers and basins. ByteStorm (red) closely follows the observed IBTrACS variability (blue) on all considered basins.
Despite the periodic sampling characterizing the test set (see Section \ref{sect:patch_gen_dataset}), ByteStorm captures the temporal variability of TCs with high fidelity, suggesting that the DL models correctly detect the frequency of TC over the multi-decadal period.

Statistical robustness is further confirmed by the high Pearson correlation values reported in Table \ref{tab:pearson_iav}, computed between the IAV of IBTrACS observations and that of each TC tracker across the test set months spanning 40 years (see Section \ref{sect:val_metrics}). In each basin, ByteStorm achieves at least the second-highest correlation, ranging from $0.66$ (NI) to $0.91$ (SI). Unlike deterministic trackers, which rely on fixed thresholds, ByteStorm's convolutional kernels adaptively classify and localize TC centers with respect to the basin of application.

\begin{figure}
    \centering
    \includegraphics[width=\textwidth]{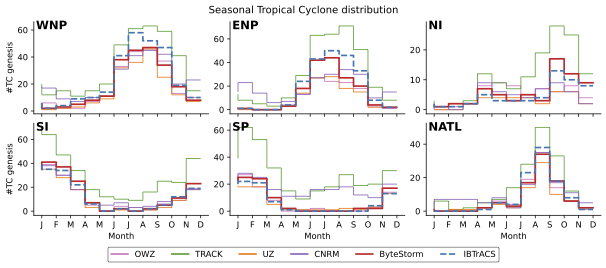}
    \caption{
    Seasonal distribution of TC genesis across the different TC formation basins over the test set period. The x-axis shows the month of occurrence, while the y-axis reports the number of TC genesis events aggregated over 40 years for each TC tracker. ByteStorm is reported as a red thicker line, while IBTrACS is represented as a dashed blue line. 
    }
    \label{fig:seas_dist}
\end{figure}

To further characterize seasonal variability, Figure \ref{fig:seas_dist} shows the distribution of TC genesis by month in each basin, aggregated over 40 years of test data. ByteStorm (red) accurately reproduces the seasonal cycle reported in IBTrACS, closely matching the timing and amplitude of TC peaks, with some exceptions like WNP and ENP basins where ByteStorm slightly under-estimates the TC frequency during July, August and September. Generally, in agreement with other trackers and consistent with the physical properties of each basin, ByteStorm shows maximum TC frequency during summer in the Northern Hemisphere (WNP, ENP, NI, NATL) and during winter in the Southern Hemisphere (SI, SP).

\subsubsection{Track Duration}
\label{sect:track_duration}

\begin{figure}
    \centering
    \includegraphics{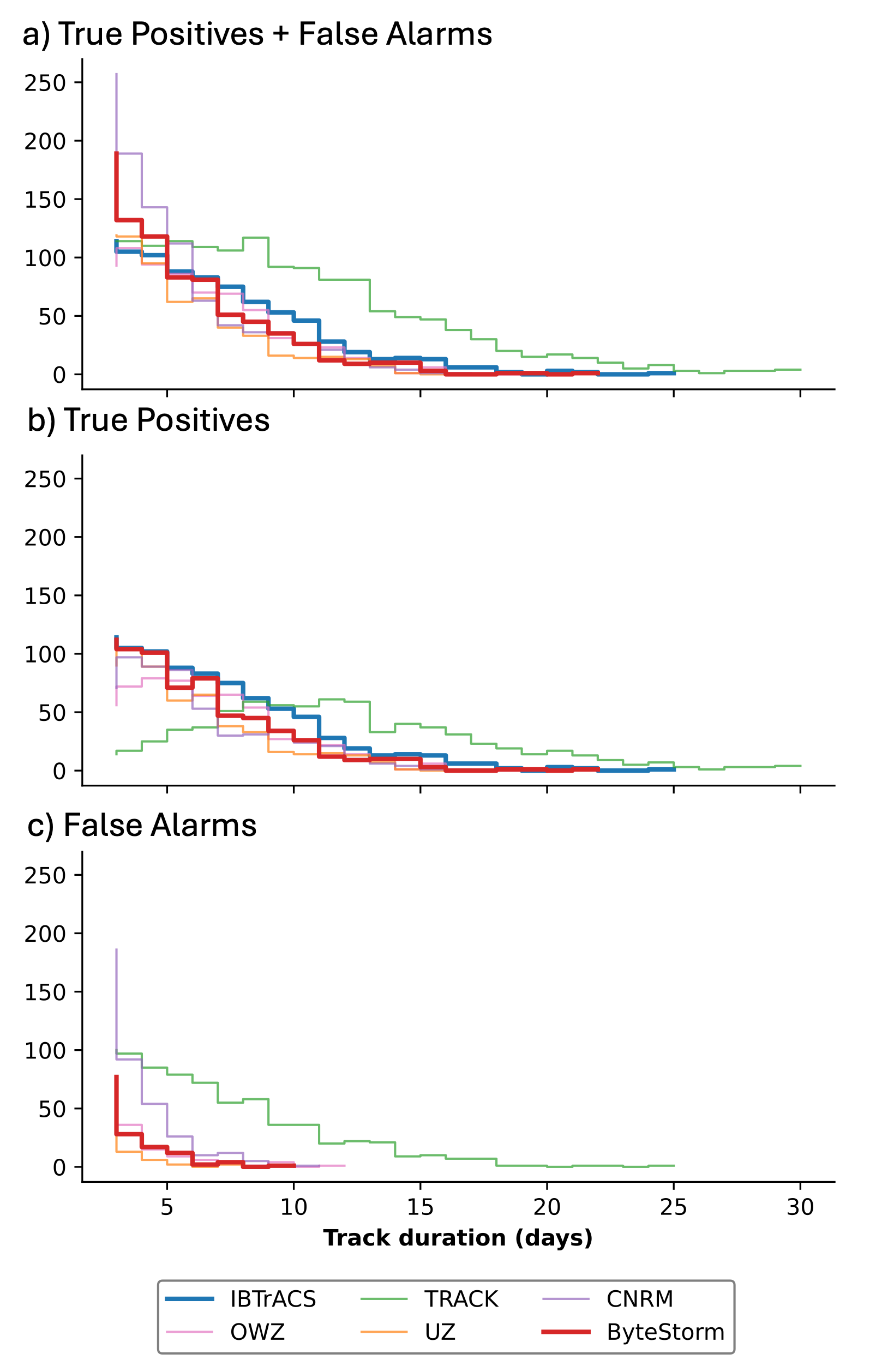}
    \caption{
    Track durations are represented as the number of TCs (y-axis) lasting a given number of days (x-axis). IBTrACS observations are shown in blue, ByteStorm in red, and TRACK, CNRM, OWZ, and UZ in green, purple, pink, and orange, respectively. Panel a) includes both TPs and FAs, while panels b) and c) show the distributions of TPs and FAs separately. 
    }
    \label{fig:track_duration}
\end{figure}

Figure \ref{fig:track_duration} (Panel a) shows the overall track duration (including True Positives - TPs - and False Alarms - FAs) produced by each tracker compared to historical observations from IBTrACS (blue). Generally, all trackers broadly reproduce the observed distribution of TC durations. In particular, UZ (orange), CNRM (purple), and OWZ (pink) closely follow the IBTrACS distribution, while TRACK (green) tends to overestimate both the number of TC days and the overall track duration, especially for medium-lasting TCs (10-15 days). Panel b), which includes only TPs, further highlights the agreement between the TC trackers and the observed TC lifetime. Consistent with Panel a), TRACK is the only tracker that systematically overestimates TC duration, producing a larger number of storms that persist longer than those reported in IBTrACS.

Although ByteStorm and CNRM reproduce track durations comparable to the observations, they generate a relatively large number of short tracks (i.e., TCs lasting $\sim 3$-$5$ days), which correspond to FAs (Panel c). In the next section we investigate both FAs and Missed tracks. 

\subsubsection{ByteStorm Missed and False Alarms tracks}
\label{sect:miss_and_fas}

\begin{figure}
    \centering
    \includegraphics[width=\textwidth]{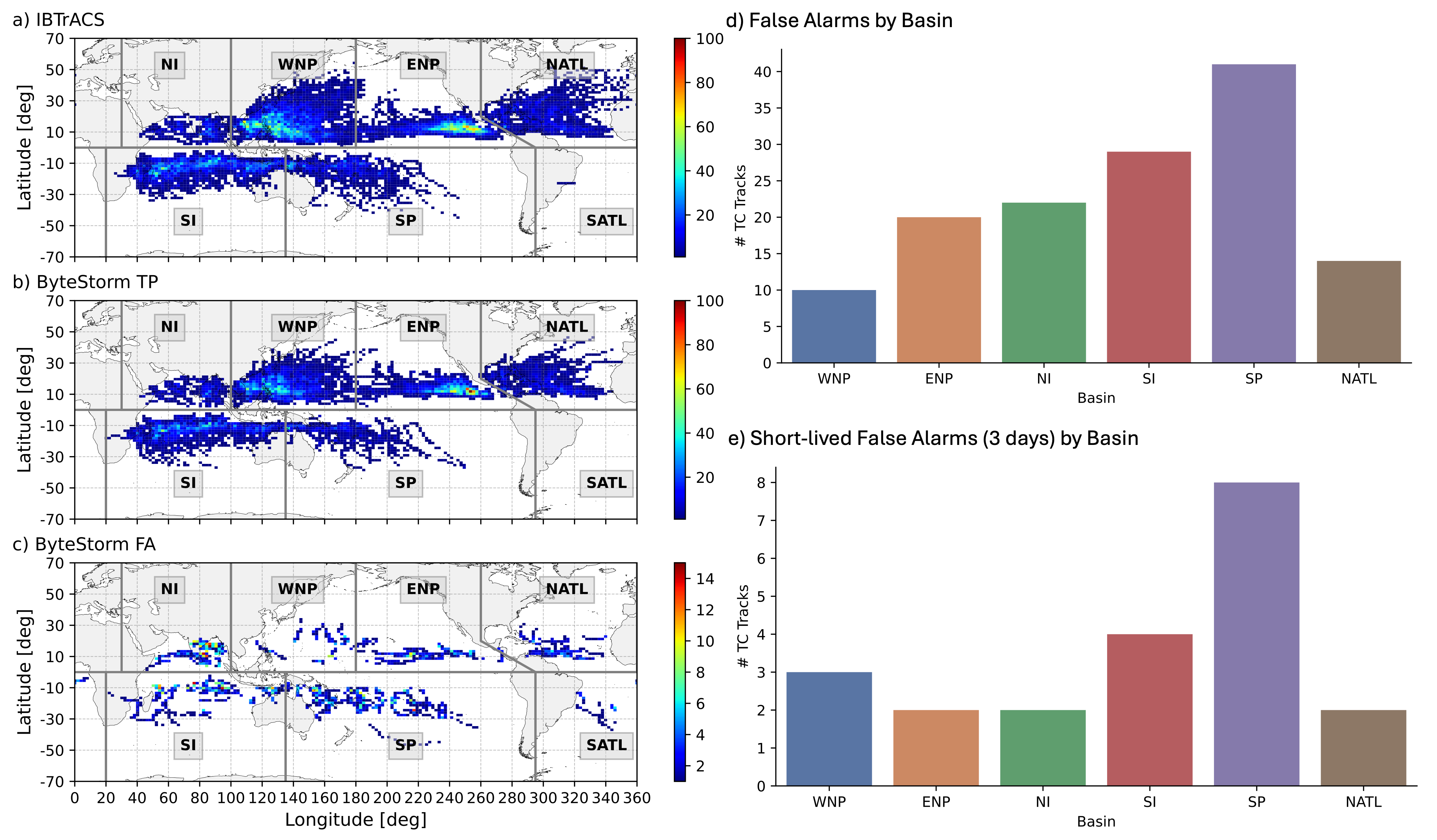}
    \caption{
    Storm track density of ByteStorm over the globe on a $2^{\circ} \times 2^{\circ}$ grid. Panel a) shows IBTrACS track density, panel b) shows ByteStorm's correctly matched TC tracks, whereas panel c) depicts TC tracks not corresponding to any historical track. 
    Panel d) presents the distribution of all ByteStorm FAs across the different basins, Panel e) focuses specifically on FAs with a duration of exactly three days. 
    }
    \label{fig:density_tp_and_fa}
\end{figure}

To further analyze Missed TCs and FAs, we perform a spatial assessment of FA geographical distribution. Figure \ref{fig:density_tp_and_fa} shows global storm density from IBTrACS (Panel a), ByteStorm TPs (Panel b) and FAs (Panel c), computed on $2^{\circ} \times 2^{\circ}$ grid cells. Comparing Panels a) and b), we observe missed detections above $40^{\circ}$ latitude in WNP and NATL, likely due to structural changes of TCs shape as they transition to the extra-tropical phase. 

Panel c) shows that FAs are largely confined to the Tropics (i.e., $[-30^{\circ}, 30^{\circ}]$ latitude). Examining the environmental conditions around FAs within a $8^{\circ} \times 8^{\circ}$ radius, we find maximum RV850 of $\sim$ $2$-$3$ $ \times 10^{-4} s^{-1}$ and minimum MSLP of $\sim$ 1000-1003 $hPa$, consistent with weak tropical disturbances. Therefore FAs may correspond to short-lived or weak systems not included in IBTrACS due to duration or intensity below the threshold. 

Moreover, Figure \ref{fig:density_tp_and_fa} panels d) and e) show the distribution of all FAs (Panel d) and specifically short-lived FAs lasting exactly three days (Panel e) across the different basins. Panel d) indicates that most FAs originate in the Southern Hemisphere (SP and SI), followed by NI and ENP. As shown in Panel e), the SP basin exhibits the largest number of three-day-long TC tracks detected by ByteStorm. This suggests a higher sensitivity of the model towards disturbances that typically characterize this region \citep{vincent1994}. 

Resulting from aforementioned analyses, we can draw some considerations. First, high throughput of FAs aligns with our training setup, which comprises all available storm types, including disturbances (Section \ref{sect:ibtracs_filtering}). Consequently, while ByteStorm effectively captures weak systems, this heightened sensitivity also leads to the detection of spurious weak disturbances. 
Second, a consistent amount of FAs is represented by short-lived tracks. Although the presence of such TCs increases the FAR (see Section \ref{sect:pod_far}), it may also indicate that the DL models capture the full evolution of these small perturbations. We provide examples of the evolution of some FAs in Appendix \ref{sect:app_evo_fa}.
This interpretation is consistent with the results discussed earlier for POD, FAR, and Track Duration, where ByteStorm appears particularly sensitive to weather atmospheric disturbances.

\subsubsection{Track Smoothness}
\label{sect:track_smoothness}

\begin{figure}
    \centering
    \includegraphics[width=\textwidth]{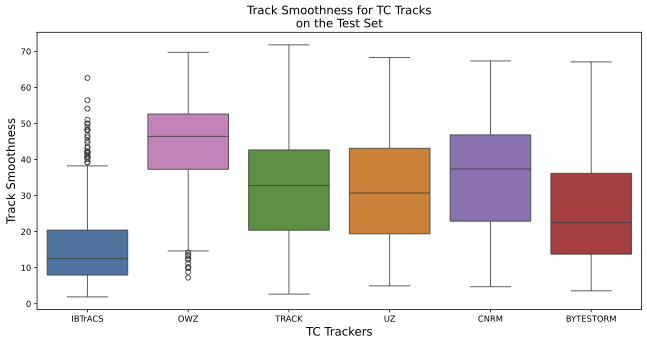}
    \caption{
    Track Smoothness box-plots comparing IBTrACS (blue, left) with deterministic trackers (pink, green, orange, and purple) and ByteStorm (red, right). For each TC track, the Track Smoothness is computed as the standard deviation of the change in the bearing angle between consecutive detections along the track. The resulting sets of smoothness scores is then summarized using box-plots to provide descriptive statistics of the distributions.
    }
    \label{fig:track_smoothness}
\end{figure}

In Figure \ref{fig:track_smoothness}, we report the Track Smoothness of IBTrACS, Bytestorm, and the deterministic trackers. Smoothness scores are computed following the methodology described in Section \ref{sect:val_metrics}. We report the resulting distributions over the test set as box-plots in Figure \ref{fig:track_smoothness}. Each box-plot represents the range between the $1^{st}$ and $3^{rd}$ quartiles, while the horizontal line within the box denotes the median of the distribution. 

ByteStorm is the TC tracker that produces substantially smoother TC tracks than the other trackers, resulting in values closer to the IBTrACS reference. Specifically, ByteStorm exhibits an Inter-Quartile Range (IQR) of $[13.74, 36.16]$ with a median in $22.49$, partially overlapping with the IBTrACS distribution, whose IQR spans $[7.94, 20.42]$.

We attribute these results to two main factors. First, the integration of the BYTE algorithm allows the tracker to handle temporary TC detection failures. When the DL model fails to detect a TC center at a given 6-hour timestep, BYTE uses Kalman Filter to provide a consistent estimate of the next TC center position based on previous track state. This mechanism effectively smooths the trajectory and prevents abrupt jumps when the tracker resumes the track after a missed detection.

Second, the DL models are threshold-agnostic, meaning that they learn to identify morphological patterns through convolutional kernels, enabling the detection of spatially consistent TC centers that adapt to the climatological characteristics of different basins. 

Overall, these results shows that the combined use of DL models and the BYTE algorithm produces smooth TC tracks and improves temporal consistency between consecutive time-steps, although the DL models themselves are not explicitly trained to account for temporal dynamics.

\subsection{Test Cases: Results on long-lived TC tracks}
\label{sect:results_longest_tracks}

In this Section, we analyze the tracks of five long-lived TCs as representative test cases. These storms are selected from the test set by sampling events from different basins and spanning a broad temporal range. The selected TCs are Owen (August 1989, WNP), John (August 1994, WNP), Dora (August 1999, ENP), Nicholas (October 2003, NATL) and Gafilo (March 2004, SI).
In this section we focus on TCs Gafilo and John, while the remaining cases are presented in Appendix \ref{sect:appendix_b}, together with general information about the life-cycle of the selected storms.

\begin{figure}
    \centering
    \includegraphics[width=\textwidth]{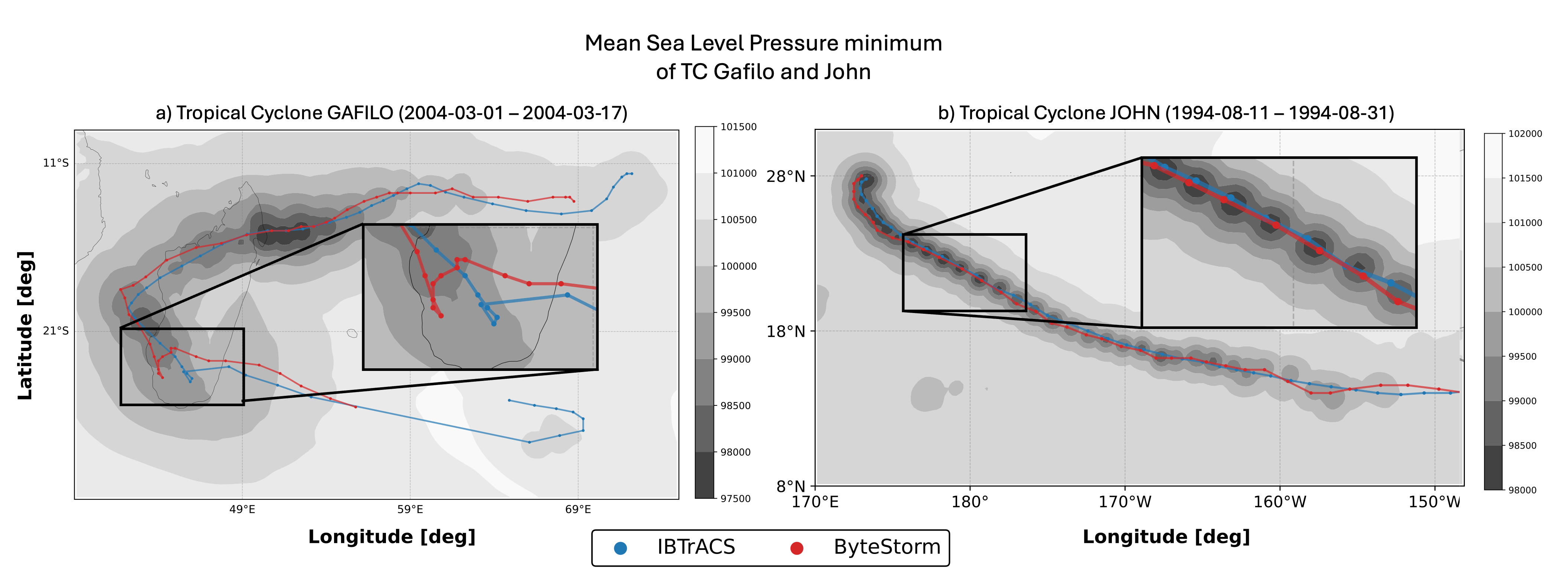}
    \caption{
    TC trackes for Gafilo (Panel a) and John (Panel b) are shown in the figure. IBTrACS historical TC positions are reported in blue, while tracks detected by ByteStorm are shown in red. The tracks are overlaid on the minimum MSLP computed across the entire observed TC lifetime, highlighting the low-pressure centers associated with the storm throughout its evolution. 
    }
    \label{fig:msl_gafilo_john}
\end{figure}

Figure \ref{fig:msl_gafilo_john} shows the observed tracks of TC Gafilo (Panel a) and TC John (Panel b) from IBTrACS (blue), alongside the corresponding detections by ByteStorm (red). Both the observed and detected tracks are overlaid on maps of ERA5 MSLP, obtained by computing the minimum MSLP at each time-step along the TC track. This visualization highlights the associated low-pressure centers, facilitating the spatial localization of the storms. \\

\noindent \textit{Tropical Cyclone Gafilo}
\label{sect:tc_gafilo}

During the early stages of TC Gafilo formation (Panel a), top right), ERA5 data shows only a weak and poorly defined circular pattern in the MSLP field. During this phase, ByteStorm does not accurately follow Gafilo's track. 
As the storm becomes more organized and intensifies (Panel a), center), the MSLP minima appear as clearly defined darker regions along the TC track, and ByteStorm's error becomes very low. Finally, Gafilo exhibits a $\gamma$-shaped trajectory, highlighted in the zoomed portion of Panel a). ByteStorm captures a similar motion, although with a slight spatial offset. At this stage, most the TC trackers fail to correctly reproduce the observed $\gamma$-shaped path, as shown in Appendix Figure \ref{fig:msl_5_tracks}, Panel c).

\noindent\textit{Tropical Cyclone John}
\label{sect:tc_john}

TC John has been considered one of the most intense cyclonic systems ever recorded in the Pacific. In Panel b) of Figure \ref{fig:msl_gafilo_john}, we show TC John during the phase when its intensity was increasing and the system was gaining energy. The map clearly highlights the low-pressure minima, visible as dark regions centered on the historical TC positions from IBTrACS (blue). ByteStorm track (red) closely match the observed TC centers, as illustrated in the zoomed portion of the figure. As the storm begins to weaken (left side Panel b)), ByteStorm detects the TC centers with lower accuracy, showing a gradual drift from the observed TC centers.

\subsection{Extrapolation on 2020-2022: ERA5 and JRA-3Q reanalysis}
\label{sect:res_2020_2022}

We now assess the robustness of the proposed approach on more recent years and across different datasets, thereby providing insights into the broader applicability of ByteStorm.
Exploiting several of the metrics introduced earlier, we evaluate ByteStorm on both ERA5 and JRA-3Q on the same test period: 2020-2022.
We do not repeat the benchmark with deterministic TC trackers here, as Section \ref{sect:res_comp_trackers} already provides a comprehensive and statistically robust evaluation. 

\begin{figure}
    \centering
    \includegraphics[width=\textwidth]{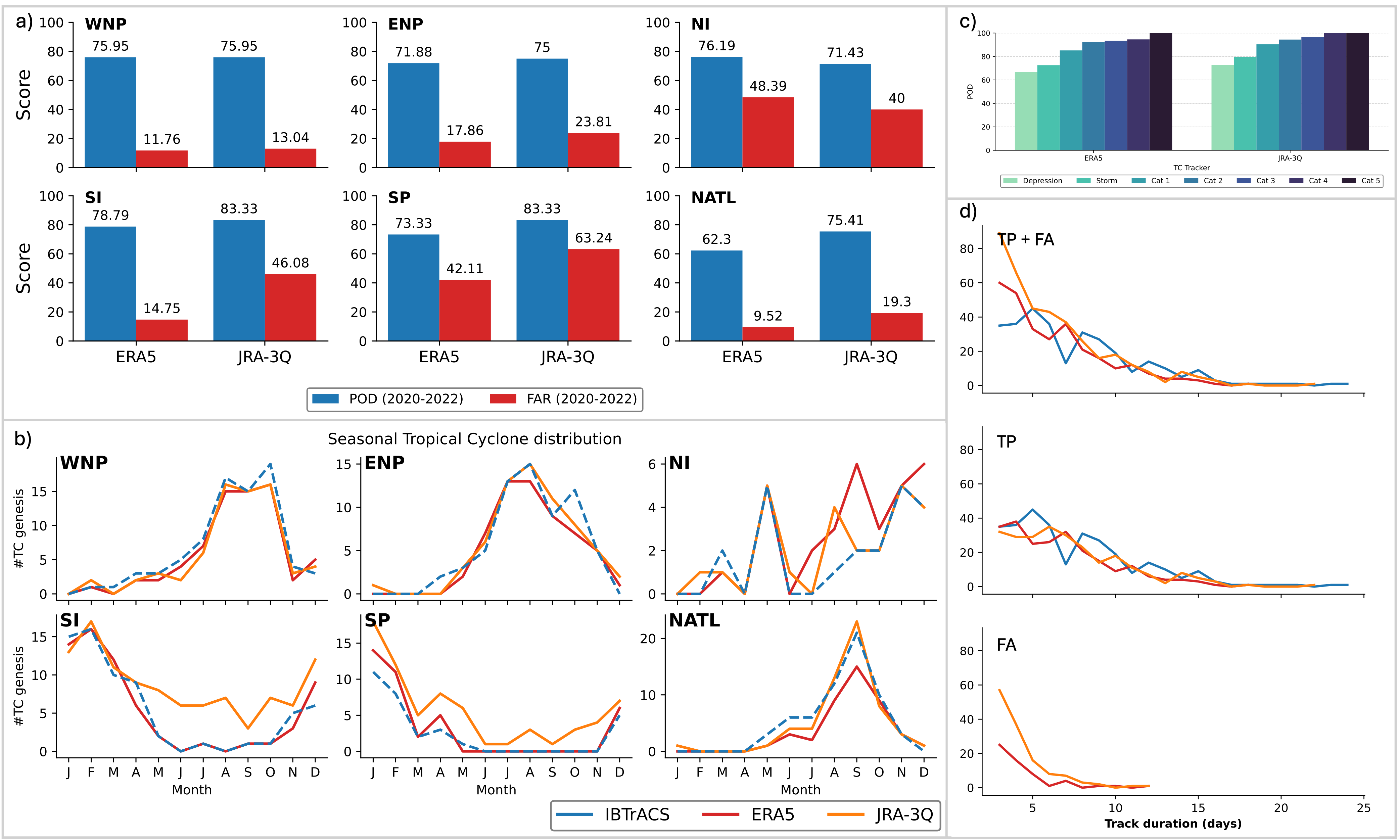}
    \caption{
    A comparison of ERA5 and JRA-3Q dataset is shown over 2020-2022 period.
    In panel a) and b) is reported POD-FAR bar-plot and seasonal tropical cyclone distribution across the considered period, respectively.
    Panel c) shows POD divided by storm category according to the Saffir-Simpson scale, whereas panel d) shows the track duration over the two datasets compared with IBTrACS ground truth.
    }
    \label{fig:era5_jra_1}
\end{figure}

Considering Figure \ref{fig:era5_jra_1}, ByteStorm's skills on ERA5 during 2020-2022 remain consistent with the results obtained for the 1980-2019 test set, indicating good temporal extrapolation performance. When applied to JRA-3Q, ByteStorm achieves similar —and in some cases higher— POD, but at the cost of an increased FAR (Panel a). For instance, in the SP basin POD and FAR are $10 \%$ and $20 \%$ higher than ERA5's ones. We observe a similar behavior in NATL basin, where both POD and FAR on JRA-3Q are $\sim10\%$ higher.

To better understand the simultaneous increase in POD and FAR, we perform additional analyses. 
First, we discretize POD according to the SSHS scale (Figure \ref{fig:era5_jra_1}, Panel c). This analysis shows that 
POD score increase on JRA-3Q compared to ERA5 for Depressions, Storms and Category 1-2 TCs. 
Second, JRA-3Q's seasonal TC distribution reveals an anomalously high number of TCs outside the typical seasonal cycle in the SI, SP and NI basins (Panel b). In contrast, in the remaining basins both ERA5 and JRA-3Q closely follow the expected seasonal TC frequency.
Third, Panel d) shows ByteStorm's track duration distributions for ERA5 (red) and JRA-3Q (orange), compared with IBTrACS observations (blue). Focusing on the FAs distribution, JRA-3Q produces a substantially larger number of short-lived tracks, consistent with the higher FAR.

Taken together, these results suggest that ByteStorm reliably extrapolates on recent ERA5 years and is more sensitive to TC detection when applied to JRA-3Q data, providing large amounts of short tracks. Latter behavior may arise either from the different representation of TCs in ERA5 and JRA-3Q or the lack of dedicated fine-tuning on JRA-3Q. A more detailed investigation would be required to fully explain the increased detection rate (including both TPs and FAs) observed with JRA-3Q; however, since this is out of this study's scopes, we leave such analysis for future work. In Section \ref{sect:app_era_jra} we provide additional evaluations for ERA5 and JRA-3Q over the 2020–2022 test period.

\subsection{Computational Performance}
\label{sect:compute_performance}

We trained the DL models (see Section \ref{sect:detection}), exploiting $4$ GPUs NVidia Volta A100 40GBs, two compute nodes of the Juno hybrid cluster - one of the High Performance Computing (HPC) systems available at the CMCC \citep{cmcc_hpcc}, and a single GPU at test time.

To evaluate computational requirements of ByteStorm, we measure the execution time and the amount of data needed to compute TC tracks in an example scenario. We run ByteStorm and UZ (based on TempestExtremes from \cite{ullrich_2021}) for 3 year of 6-hourly ERA5 data, resulting in a total of 4384 timesteps. 
Table~\ref{tab:benchmark_comparison} reports a summary of the results. Execution times represent the average time of three different runs of the same configuration, while the input size is the volume associated with the variables required by the trackers. We execute ByteStorm configured to use a single GPU for the DL models and a single CPU for the other steps, while UZ uses a single CPU.

As we can see in Table~\ref{tab:benchmark_comparison}  the amount of data needed for running ByteStorm, depending only on two variables, is around 28\% of the amount needed for running the UZ tracker (i.e., 66 GB less). Moreover, ByteStorm executes nearly $7 \times$ faster than UZ. 
This shows the inherent advantage of data-driven model designed for running on GPUs. 

A complete computing performance comparison of the different trackers implementation is out of the scope of this paper. \cite{bourdin_2022} already provides a benchmark of TC trackers' execution times, stating that except for CNRM, which takes twice the time, all the other TC trackers have similar execution time. Under this assumption, we consider the speedups observed with respect to UZ as indicative of ByteStorm's performance over traditional tracking approaches. 

\begin{table*}
    \centering
    \caption{Performance evaluation of TC trackers on 3 years of ERA5 data (4384 timesteps).}
    \label{tab:benchmark_comparison}
    \begin{tabular}{lcccc}
        \hline
        \textbf{Model} & \textbf{Execution Time (s)} & \textbf{Input size (GB)} & \textbf{Throughput (ts/s)} \\
        \hline
        ByteStorm      & 146.8     & 26.4  & 29.9     \\
        UZ (TempestExtremes) & 980.4    & 92.4  & 4.5       \\
        \hline
    \end{tabular}
\end{table*}

\section{Discussion}
\label{sect:discussion}

ByteStorm is a fully data-driven approach that combines two DL models with BYTE algorithm, requiring only two environmental predictors—RV850 and MSLP—reducing data volume and computational cost, while still learning the physical patterns associated with TC morphology. 

Benchmarking against four deterministic trackers demonstrates that ByteStorm reliably reconstructs smooth and coherent TC tracks, reproduces both seasonal and inter-annual variability, and maintains consistent detection skills across basins (Figures \ref{fig:pod_far}, \ref{fig:iav}, \ref{fig:seas_dist}). It captures track duration distributions with high fidelity, despite a slight over-representation of short tracks due to false alarms or small transient disturbances (Figure \ref{fig:track_duration}). Case studies of long-lived TCs further confirms the model’s robustness, accurately capturing cyclone paths during intense phases while producing physically realistic tracks (Figure \ref{fig:msl_gafilo_john}).

ByteStorm also offers practical advantages over deterministic methods. Its DL components can leverage GPU acceleration, enabling faster computation, and its simplified input requirements reduce storage and pre-processing costs. Additionally, the combination of DL detection with physically constrained data association method ensures smooth and temporally consistent tracks, often outperforming deterministic approaches in terms of track smoothness (Figure \ref{fig:track_smoothness}).

Limitations remain. ByteStorm does not predict TC intensity alongside position, and it generates a number of short-lived tracks that may correspond to model hallucinations or small-scale disturbances. The use of two separate CNNs for classification and localization increases training time. Additionally, training a couple of models for each hemisphere, further increase the complexity of the setup, which could be mitigated by unified architectures in future developments.

Overall, ByteStorm demonstrates that data-driven TC tracking can complement traditional deterministic methods, providing a computationally efficient, physically consistent, and robust approach for global TC monitoring. Its flexibility and minimal parameter requirements make it well suited for large-scale climate diagnostics, high-resolution reanalysis evaluation, and potential integration into operational TC tracking pipelines, while maintaining comparable skills to traditional trackers.

\section{Conclusions}
\label{sect:conclusions}

We introduced ByteStorm, a novel framework combining deep learning (DL)–based classification and localization models with BYTE \citep{zhang_2022}, a state-of-the-art computer vision Multi-Object Tracking (MOT) algorithm, to reconstruct tropical cyclone (TC) tracks in gridded environmental data. 
By linking individual TC centers into temporally coherent tracks, ByteStorm provides an end-to-end, scalable solution for storm detection and tracking.
Unlike traditional deterministic trackers, which rely on manually tuned thresholds, ByteStorm follows a fully data-driven paradigm. 
It enables quick, flexible and cheap statistical studies on TC tracks, thereby significantly broadening both the potential user base and applications. We used an earlier version, for example, as a data-driven component powering a Digital Twin application for extreme events analysis \citep{manzi26}.

Importantly, we remark that the objective of this work is not to compare deterministic and data-driven approaches; deterministic trackers remain well-established tools for TC tracking in climate datasets. Instead, ByteStorm is presented as a novel complementary and computationally efficient framework with performance comparable to existing state-of-the-art methods. It exhibits good TC track reconstruction and extrapolation skills, globally, highlighting its robustness and potential applicability on other reanalysis datasets with none or minimal fine-tuning effort (see Section \ref{sect:res_2020_2022}).
Additionally, the computational efficiency and scalability of the framework further enables both fast and reliable TC diagnostics within large climate datasets, and provides a powerful DL-based analysis tool for operational and real-time TC tracking and monitoring.

Looking ahead, different promising directions emerge to enhance the current framework. Exploring other of DL techniques or embedding physical constraints within the data-driven approach, may enable better performance in the detection/localization task, thus enhancing the overall tracker robustness. Furthermore an important direction is the development of a unified model capable of jointly performing all tasks together while also predicting additional key TC-related information such as intensity and direction. 

Finally, future work will prioritize its integration in Earth System Models as a diagnostic tool, enabling real-time TC diagnostics as a direct output of climate simulations. This will pave the way for more effective and reliable monitoring and forecasting of TCs on a global scale.

\section*{CRediT authorship contribution statement}

\noindent\textbf{Davide Donno:} Writing – original draft, Visualization, Methodology, Conceptualization, Investigation, Software, Data curation, Validation\\
\textbf{Donatello Elia:} Writing – original draft, Visualization, Methodology, Conceptualization, Investigation, Supervision, Validation\\
\textbf{Gabriele Accarino:} Writing - Review \& Editing, Methodology, Conceptualization, Data Curation, Validation\\
\textbf{Marco De Carlo:} Writing – original draft, Visualization, Investigation, Software, Validation\\
\textbf{Enrico Scoccimarro:} Writing - Review \& Editing, Conceptualization, Visualization, Validation\\
\textbf{Silvio Gualdi:} Writing - Review \& Editing, Conceptualization, Visualization, Validation\\

\section*{Funding}

This work was supported in part by the interTwin project. interTwin has received funding from Horizon Europe under grant agreement No 101058386.

\section*{Computer Code Availability}
\begin{itemize}
    \item Software name: \textit{ByteStorm}
    \item Programming language: Python
    \item Software requirements: UNIX, PyTorch, Xarray, Numpy, Pandas
    \item Hardware requirements: It is recommended to use a system equipped with one or more GPUs and at least 32 GB of RAM.
    \item Source code is available for downloading at: \url{https://github.com/CMCC-Foundation/bytestorm}
    \item Program size: 131 MB considering the provided repository. The whole dataset that we used is ~350 GB.
\end{itemize}

\section*{Data availability}
\noindent Datasets used for model development and training
\begin{itemize}
    \item ERA5 single levels dataset: \url{https://cds.climate.copernicus.eu/datasets/reanalysis-era5-single-levels?tab=overview}
    \item ERA5 pressure levels dataset: \url{https://cds.climate.copernicus.eu/datasets/reanalysis-era5-pressure-levels?tab=overview}
    \item IBTrACS dataset: \url{https://www.ncei.noaa.gov/products/international-best-track-archive}
    \item JRA-3Q dataset: \url{https://gdex.ucar.edu/datasets/d640000}
    \item Tracks for OWZ, TRACK, UZ and CNRM trackers: \url{https://zenodo.org/records/6424432}
\end{itemize}

\bibliographystyle{cas-model2-names}
\bibliography{main}

\newpage
\section{Appendix}
\label{sect:appendix}

\subsection{Hyperparameter tuning of BYTE}
\label{sect:appendix_a}

We use an optimization algorithm to select the best configuration of data-driven models and BYTE-based tracking. Specifically, fixing the bounding box size to $25$, the \textit{Track Buffer} to $1$ and varying both the \textit{Match} and \textit{Track Threshold}, we perform a first grid-search. This preliminary operation aims at finding the best candidates for \textit{Match-} and \textit{Track-Threshold} while, at the same time, avoiding the high computational cost of an entire grid search. Within the pool of candidates, we select and refer to as the \textit{Baseline} the TC tracker having \textit{Match Threshold} = $0.8$ and \textit{Track Threshold} = $0.7$. We select this configuration as it produces an acceptable POD-FAR trade-off (i.e. high POD and low FAR, comparable to the other trackers) and good Pearson correlation on all the basins.

After this phase, we then vary the \textit{Bounding Box Size} in the set $BB = \{15, 21, 25, 31, 35\}$ and the \textit{Track Buffer} in $TB=\{1, 2, 3, 4\}$. Since the spatial resolution of the ERA5 climatic predictors is $0.25^{\circ} \times 0.25^{\circ}$, we choose the bounding box size lower bound (15) and upper bound (35) in order to have physically meaningful DL-trackers, thus avoiding TC tracks with too far detections (i.e., above $400$ km). On the other hand, we set the \textit{Track Buffer} to a maximum of $4$ time-steps, similarly to \cite{ullrich_2021} tracker. The total number of possible DL-Trackers is $20$ in this phase.

We augment the pool of candidates including some physical constraints, applied one at a time: i) removing TCs originating on land, ii) removing TCs originating above $30^{\circ}$ Latitude (\cite{hodges_2017}) and iii) removing TCs originating above $50^{\circ}$ Latitude (\cite{ullrich_2021}). In addition to the aforementioned $3$ post-processing, we add the combination of i)-ii) and i)-iii). Therefore, we reach a total of $120$ different DL-trackers candidates.

Among the resulting pool of DL-Trackers, we select the best Tracker according to a simple weighted multi-objective optimization algorithm based on the Pareto Frontier. The targeted metrics to be optimized are: POD, FAR, and Pearson Correlation on the whole domain considered in the study. The assigned weight \textit{w}, are $w_{POD} = 0.5$, $w_{FAR} = 0.4$ and $w_{IAV} = 0.1$ respectively. The resulting DL tracker is the one with bounding box size equal to $31$, with track duration of $2$ steps and excluding TCs originating above $30^{\circ}$ Latitude, with global skills: $POD = 78.08 \%$, $FAR = 17.78 \%$ , and $r_{IAV} = 0.83$. Hence, we select this configuration to be the one we use in this work. We use the same setup of BYTE for all the experiments reported in the paper on both ERA5 and JRA-3Q without any change.

\subsection{Additional results of ByteStorm}
\label{sect:appendix_b}

Aiming at assessing the skills of ByteStorm and extending the results already presented, we report additional details and results in this section. Specifically, we detail the latitude-longitude distribution between IBTrACS observations and the predicted TC centers, and we provide additional details regarding the 5 longest TC tracks available in the test set. Moreover, in Section \ref{sect:app_era_jra} and \ref{sect:app_evo_fa} we further provide results on the extrapolation skills and the evolution of FAs, respectively.

\subsubsection{Latitude and Longitude distribution}
\label{sect:latlon_dist}

\begin{figure}
    \centering
    \includegraphics[width=\textwidth]{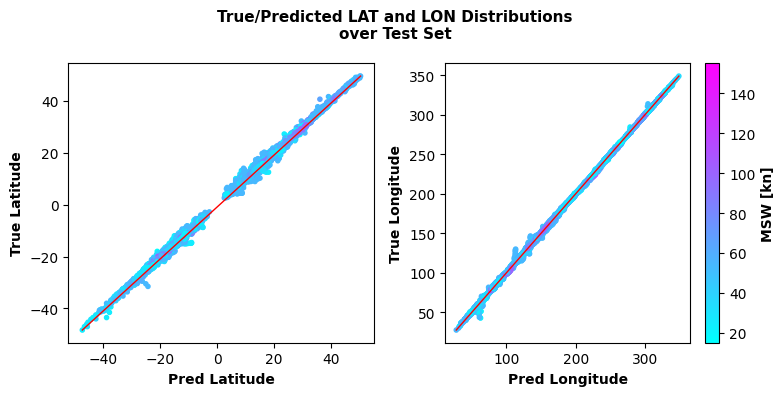}
    \caption{
    Distribution of Predicted-True Latitudes (left) and Longitudes (right). Red bisector lines represent the perfect matching between predicted and true positions, while each colored dot represent the predicted TC eye position with respect to the correct one. The color of the dots represents the Maximum Sustained Wind of the TC observation.
    }
    \label{fig:lat_lon_dist}
\end{figure}

Regarding the localization accuracy, in Figure \ref{fig:lat_lon_dist} we show the predicted and true Latitude (left panel) and Longitude (right panel) distributions over the test set. Red bisector line represents the theoretical perfect alignment between true and predicted TC geographical coordinates. The side color bar reports the MSW of the TCs matched between IBTrACS and ByteStorm.
As we can see in the image, longitudes are mostly distributed around the bisector line, while the lower latitudes have higher variability in terms of positional accuracy. Bytestorm is biased towards underestimating the real TC latitude, as the dots in Figure \ref{fig:lat_lon_dist} are mostly located above the red bisector. Additionally, it appears that stronger TCs (i.e., characterized by high MSW) report high positional precision, meaning that ByteStorm is more accurate when the TCs are characterized by well-formed wind patterns. 

\subsubsection{Additional results on the 5 Longest TC Tracks}
\label{sect:app_c_5_longest_tcs}

In this sub-section, after a brief description of the 5 longest TC tracks considered in this work, additional information is provided to better highlight the skills of ByteStorm.

\begin{figure}
    \centering
    \includegraphics[width=\textwidth]{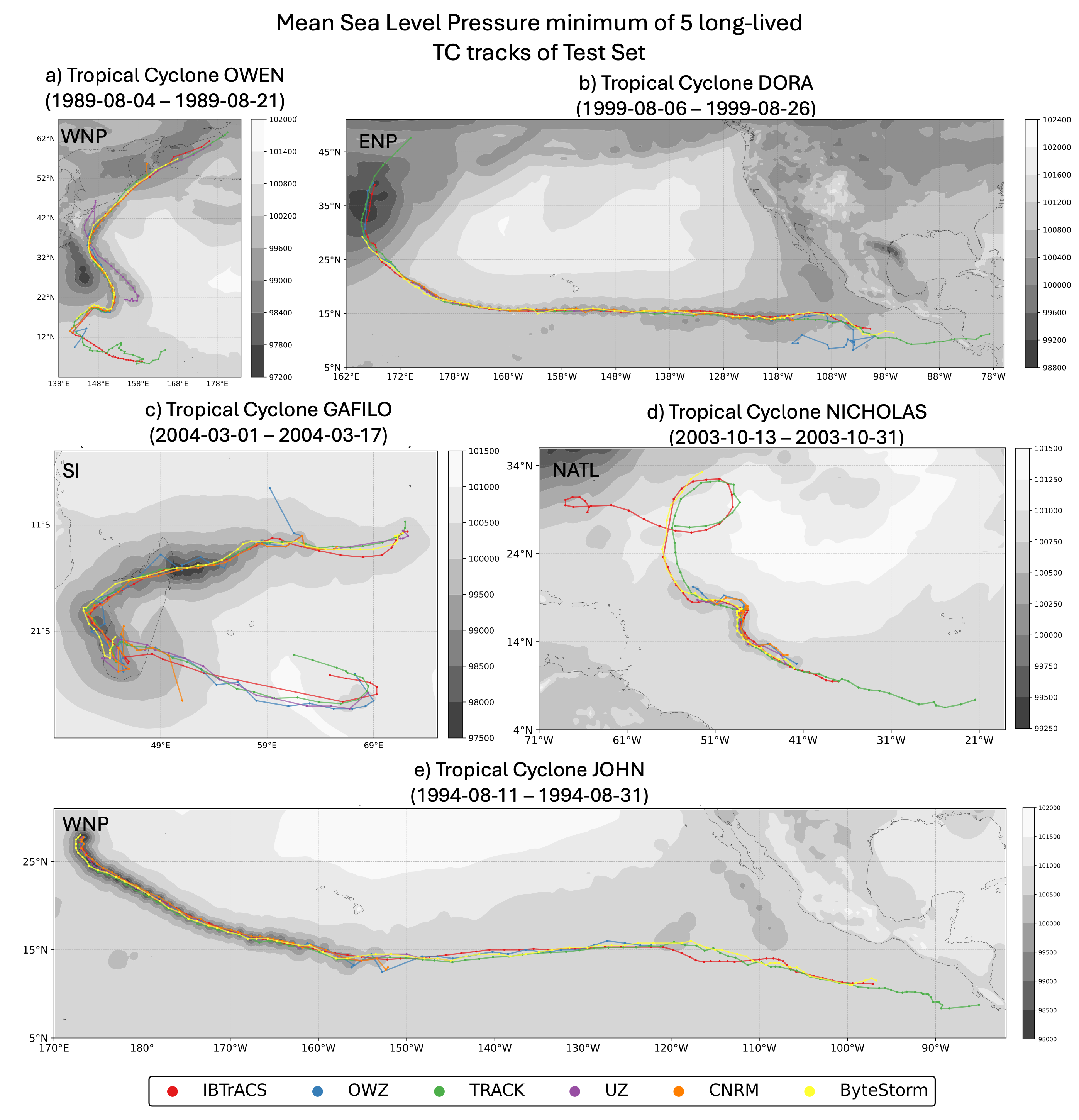}
    \caption{
    IBTrACS (in red) and detected TC centers (other colors) of 5 long-lived TC tracks of the Test Set. The TCs reported in the figure, according to the alphabetical order of the panels, are: a) Owen, b) Dora, c) Gafilo, d) Nicholas and e) John. Detected TC Tracks are overlaid on the MSLP minimum computed over the entire TC track at 6-hourly time-steps, to highlight the local low-pressure basins. 
    }
    \label{fig:msl_5_tracks}
\end{figure}

\begin{itemize}
    \item \textbf{Tropical Cyclone Owen} (1989): TC Owen started as a monsoon and slowly evolved to storm due to the proximity to TC Nancy. TC Nancy influenced TC Owen's track, forcing it to follow an unusual south-eastward trajectory. In Figure \ref{fig:msl_5_tracks} panel a) we can clearly recognize TC Nancy presence as darker MSLP regions nearby real the TC track. UZ Tracker misleads to detect TC Owen as it follows TC Nancy track (purple) \citep{report_1989}.
    \item \textbf{Tropical Cyclone John} (1994): TC John was considered one of the most intense Hurricanes ever recorded in the Central Pacific, as its MSW reached $150 kn$. It caused several heavy rains phenomena and minor floodings over some islands in the Hawaii. The damage caused by the TC was estimated to about $15$ million dollars \citep{report_1994}.
    \item \textbf{Tropical Cyclone Dora} (1999): Dora's cyclonic system during its lifetime originated about 300 miles south of Acapulco, in Mexico. As it moved west-ward it gained energy and was classified as a TS. It gained and lost energy as it moved during its path, causing small to no damages and light rain on Hawaii islands. The satellite imagery highlighted that Dora TC had a well developed TC eye \citep{report_1999}.
    \item \textbf{Tropical Cyclone Nicholas} (2003): TC Nicholas developed on 9 October starting from a tropical wave that moved westward from African coasts. It built up from a broad low pressure area and slowly gained strength and organization. The TC slowly moved west-northwestward, strengthening until it became a recognized TC, reaching MSW of 60 knots. It later weakened, until it became a depression on 23 October. Its peculiarity is the whirl-shaped trajectory (Figure \ref{fig:msl_5_tracks}, Panel d in red). \citep{report_2003}
    \item \textbf{Tropical Cyclone Gafilo} (2004): Gafilo started developing on 1st March 2004 (top right, Figure \ref{fig:msl_gafilo_john}). It continuously gained energy, until it became a powerful Category 5 cyclone whose estimated winds were 160 mph. It struck the north-east coast of Madagascar early on March 7th. It was termed as the most intense TC to hit Madagascar over the last 10 years, killing hundreds of people and damaging more than 200 000 home buildings \citep{report_2004}.
\end{itemize}

In Figure \ref{fig:msl_5_tracks} the 5 longest TC tracks are reported. Each TC tracker, as well as the IBTrACS ground truth (in red), are displayed during the TCs lifetime. To highlight the basins of MSLP minimum along the TC tracks, the sub-plots are overlaid on the minimum MSLP computed over all the 6-hourly time-steps belonging to the TC track. Aided by BYTE method, ByteStorm (in yellow) smoothly follows the historical observations for each of the considered cyclones. In contrast to other TC trackers such as CNRM (orange) and OWZ (purple), ByteStorm visually provides the smoothest tracks, as already explained in section \ref{sect:track_smoothness}. While TRACK Tracker (green) over-estimates the length of most of the TCs, our TC Tracker is closer to the real TC lifetime.

\subsubsection{Additional results on ERA5 and JRA-3Q Extrapolation}
\label{sect:app_era_jra}

\begin{figure}
    \centering
    \includegraphics[width=\textwidth]{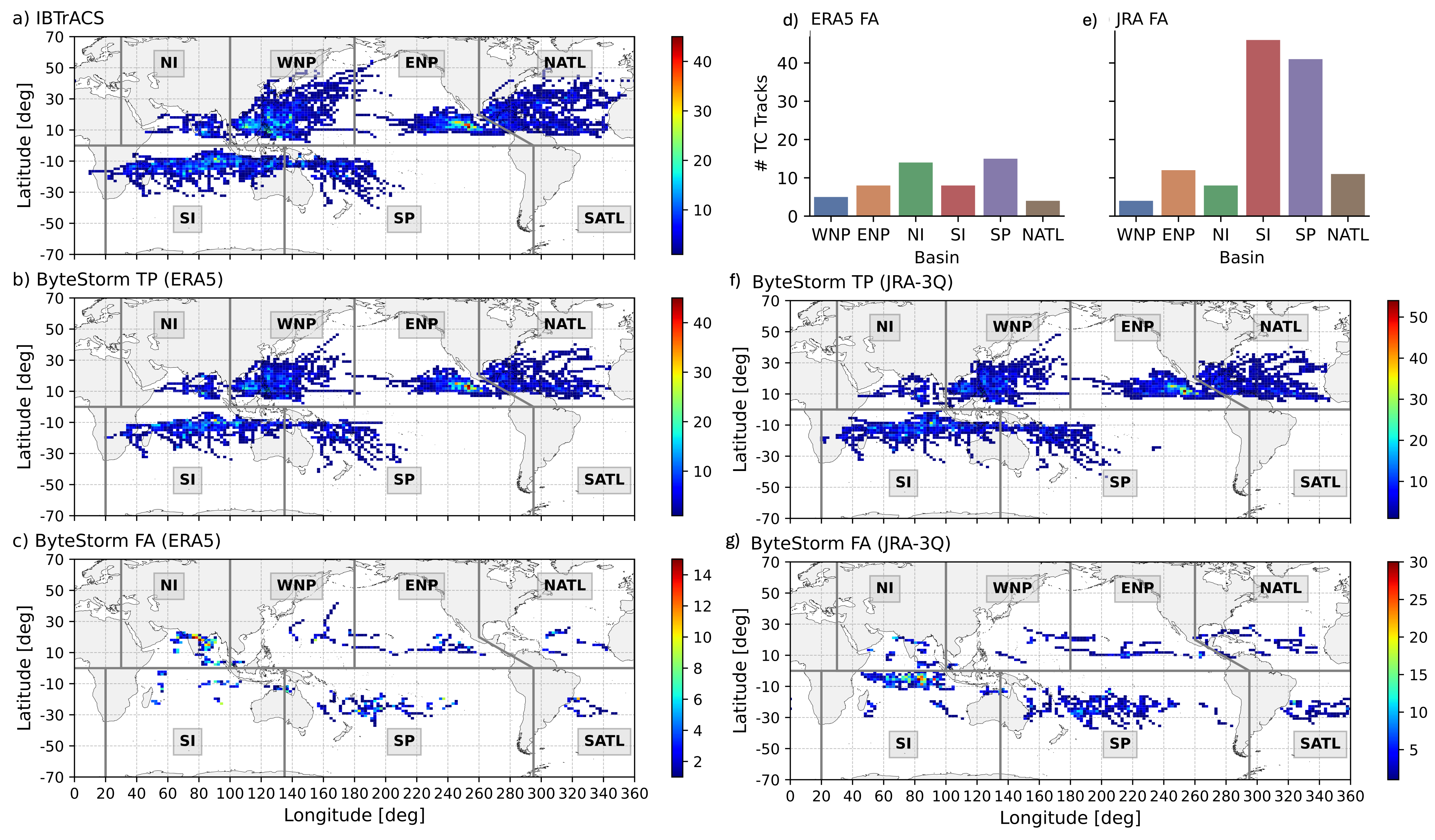}
    \caption{
    A comparison of ERA5 and JRA-3Q dataset is shown over 2020-2022 period.
    Panel a) contains IBTrACS ground truth track distribution over the test set. Panel b) and c) contain True Positives (TPs) and False Alarms (FAs) on ERA5, respectively. Panel f) and g) contain TPs and FAs on JRA-3Q, respectively. Panels d) and e) report ERA5 and JRA-3Q FAs bar-plot over each basin.
    }
    \label{fig:era5_jra_2}
\end{figure}

In this Section we extend the evaluations of Section \ref{sect:res_2020_2022}. Figure \ref{fig:era5_jra_2} contains the track density plots of IBTrACS (Panel a) and ByteStorm's TPs and FAs computed over ERA5 and JRA-3Q (Panels b and c, Panels f and g, respectively), during the 2020-2022 test set. Along with the density plots, Panels d) and e) depict the distribution of FA divided by basin over the two datasets.

Concerning ERA5, despite showing similar behavior regarding ET tracks on ENP and NATL basins (Panel b), ByteStorm is overall able to extrapolate on recent years, providing balanced number of FAs among the formation basins (Panels c and d) and substantially similar TC tracks to the IBTrACS observations (Panel a). 

Regarding JRA-3Q dataset, the figure confirms the behavior already observed in Section \ref{sect:res_2020_2022}, as ByteStorm seems to be more sensitive to TC detections on this dataset. Panel f) highlights a higher density (with respect to ERA5) in the correctly matched TC tracks, while Panel g) and e) show high number of FAs, particularly critical in SI and SP basins. 
This anomalous behavior should be better investigated from several perspectives. On one side, we should investigate the different representation of TCs between ERA5 and JRA-3Q as it may affect ByteStorm's ability to detect TCs. On the other side, we should inspect whether ByteStorm is overfitting on ERA5 data and/or a finetuning is needed to reduce JRA-3Q's FAR.

\subsubsection{Evolution of ByteStorm False Alarms}
\label{sect:app_evo_fa}

\begin{figure}
    \centering
    \includegraphics[width=\textwidth]{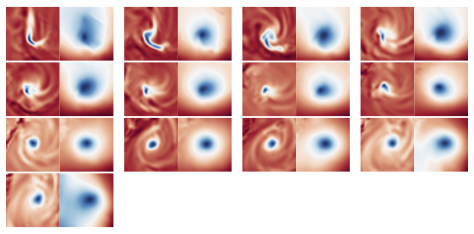}
    \caption{
    Evolution of ByteStorm's false alarm $\#1$. This false alarm evolved in LAT[-38.0, -24.25], LON[154.5, 163.75] during its lifetime (from 1990-01-24 06:00 to 1990-01-27 06:00).
    Grid layout contains (in the order) the 6-hourly timesteps centered on the TC center detected by ByteStorm. Each grid cell contains RV850 (left cell panels) and MSLP (right cell panels).
    }
    \label{fig:evo_fa_1}
\end{figure}

\begin{figure}
    \centering
    \includegraphics[width=\textwidth]{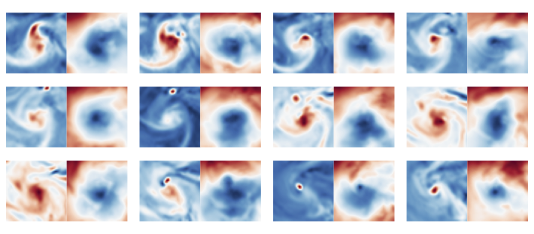}
    \caption{
    Evolution of ByteStorm's false alarm $\#2$. This false alarm evolved in LAT[7.75, 9.0], LON[153.5, 158.75] during its lifetime (from 1990-11-09 12:00 to 1990-11-12 06:00).
    Grid layout contains (in the order) the 6-hourly timesteps centered on the TC center detected by ByteStorm. Each grid cell contains RV850 (left cell panels) and MSLP (right cell panels).
    }
    \label{fig:evo_fa_2}
\end{figure}

\begin{figure}
    \centering
    \includegraphics[width=\textwidth]{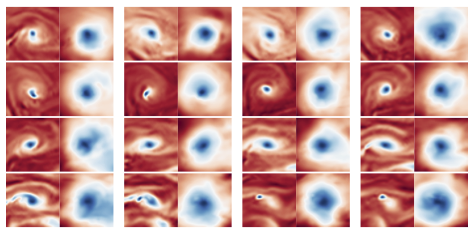}
    \caption{
    Evolution of ByteStorm's false alarm $\#3$. This false alarm evolved in LAT[-9.25, -7.5], LON[89.25, 92.25]) during its lifetime (from 2011-04-01 00:00 to 2011-04-05 00:00).
    Grid layout contains (in the order) the 6-hourly timesteps centered on the TC center detected by ByteStorm. Each grid cell contains RV850 (left cell panels) and MSLP (right cell panels).
    }
    \label{fig:evo_fa_3}
\end{figure}

Figures \ref{fig:evo_fa_1}, \ref{fig:evo_fa_2} and \ref{fig:evo_fa_3} represent the evolution of some False Alarms (FAs) detected by ByteStorm. Each of the three plots display a grid containing the 6-hourly timestep evolution of the false alarm. Sub-panels within grid cells contain RV850 (left panels) and MSLP (right panels) centered on the detected TC center.

By analyzing more in detail Figure \ref{fig:evo_fa_1} and \ref{fig:evo_fa_3}, ByteStorm is detecting the basin of minimum within MSLP alongside with RV850 radial RV850 patterns, therefore these FA can be interpreted as organized storms lasted only few days. These two FAs evolve in the southern hemisphere. 

In the case of Figure \ref{fig:evo_fa_2}, instead, ByteStorm detects a wide MSLP basin minima, with asymmetric radial patterns. Therefore, it can be interpreted as a small disturbance characterized by slow winds.

Clearly, the three FAs represent examples of ByteStorm's ability of capturing weak-entity phenomena that were not recorder in IBTrACS as they did not meet intensity and/or duration requirements.

\end{document}